\documentclass{rescience}
\usepackage{hyperref}       %
\usepackage{url}            %
\usepackage{booktabs}       %
\usepackage{amsfonts}       %
\usepackage{nicefrac}       %
\usepackage{microtype}      %
\usepackage{multirow}
\usepackage{graphicx}
\usepackage{colortbl}
\usepackage{amsmath}
\usepackage{float}
\usepackage{booktabs}

\usepackage{caption}
\usepackage{subcaption}
\usepackage{xfrac}
\usepackage{enumitem}
 
\begin{document}
\begin{tikzpicture}[remember picture, overlay]
  \node [shape=rectangle, draw=darkred, fill=darkred, yshift=-34mm,
        anchor=north west, minimum width=3.75cm, minimum height=10mm]
        at (current page.north west) {};
  \node [text=white, anchor=center, yshift=-39mm, xshift=1.875cm]
         at (current page.north west) {\small \SpaceRoboto RESCIENCE C};
\end{tikzpicture}

{\let\newpage\relax\maketitle} \maketitle

\marginnote{
  \footnotesize \sffamily
  \textbf{Edited by}\\
  \ifdefempty{\editorNAME}{\textcolor{darkgray}{(Editor)}}
             {\editorNAME\ifdefempty{\editorORCID}{}{$^{\orcid{\editorORCID}}$}}\\
  ~\\
  \ifdefempty{\reviewerINAME}{}
  {
  \textbf{Reviewed by}\\
  \ifdefempty{\reviewerINAME}{\textcolor{darkgray}{}}
             {\reviewerINAME\ifdefempty{\reviewerIORCID}{}{$^{\orcid{\reviewerIORCID}}$}\\}
  \ifdefempty{\reviewerIINAME}{\textcolor{darkgray}{}}
             {\reviewerIINAME\ifdefempty{\reviewerIIORCID}{}{$^{\orcid{\reviewerIIORCID}}$}\\}
  ~\\
  }
  \textbf{Received}\\
  \ifdefempty{\dateRECEIVED}{---}{\dateRECEIVED}\\
  ~\\
  \textbf{Published}\\
  \ifdefempty{\datePUBLISHED}{---}{\datePUBLISHED}\\
  ~\\
  \textbf{DOI}\\
  \ifdefempty{\articleDOI}{---}{\articleDOI}
}

\newcommand{\container}[1]{\def\@container{#1}}
\begin{container}
  \afterpage {
    \begin{statement}
      \scriptsize \sffamily
      \hrule \vskip .5em
      Copyright © \articleYEAR~\authorsABBRV,
      released under a Creative Commons Attribution 4.0 International license.
      
      Correspondence should be addressed to
      \contactNAME~(\href{mailto:\contactEMAIL}{\contactEMAIL})
  
      The authors have declared that no competing interests exist.

      \ifdefempty{\codeURL}{}
      {Code is available at
      \href{\codeURL}{\detokenize\expandafter{\codeURL}}\ifdefempty{\codeDOI}{.}{ -- DOI \doi{\codeDOI}.}\ifdefempty{\codeSWH}{.}{ -- SWH  \href{https://archive.softwareheritage.org/\codeSWH/}{\detokenize\expandafter{\codeSWH}}.}}

      \ifdefempty{\dataURL}{}
      {Data is available at
      \href{\dataURL}{\detokenize\expandafter{\dataURL}}\ifdefempty{\dataDOI}{.}{ -- DOI \doi{\dataDOI}.}}

      \ifdefempty{\reviewURL}{}
     {Open peer review is available at \href{\reviewURL}{\detokenize\expandafter{\reviewURL}}.}
    \end{statement}
  }
\end{container}

\section*{\centering Reproducibility Summary}

\subsubsection*{Scope of Reproducibility}

We examine the main claims of the original paper \citep{PureNoise}, which states that in an image classification task with imbalanced training data, (i) using pure noise to augment minority-class images encourages generalization by improving minority-class accuracy. This method is paired with (ii) a new batch normalization layer that normalizes noise images using affine parameters learned from natural images, which improves the model's performance. Moreover, (iii) this improvement is robust to varying levels of data augmentation. Finally, the authors propose that (iv) adding pure noise images can improve classification even on balanced training data.

\subsubsection*{Methodology}

We implemented the training pipeline from the description of the paper using PyTorch and integrated authors' code snippets for sampling pure noise images and batch normalizing noise and natural images separately. All of our experiments were run on a machine from a cloud computing service with one NVIDIA RTX A5000 Graphics Card and had a total computational time of approximately 432 GPU hours.

\subsubsection*{Results}

We reproduced the main claims that (i) oversampling with pure noise improves generalization by improving the minority-class accuracy, (ii) the proposed batch normalization (BN) method outperforms baselines, (iii) and this improvement is robust across data augmentations. Our results also support that (iv) adding pure noise images can improve classification on balanced training data. However, additional experiments suggest that the performance improvement from OPeN may be more orthogonal to the improvement caused by a bigger network or more complex data augmentation.

\subsubsection*{What was easy}

The code snippet in the original paper was thoroughly documented and was easy to use. The authors also clearly documented most of the hyperparameters
that were used in the main experiments.

\subsubsection*{What was difficult}

 The repo linked in the original paper was not populated yet. As a result, we had to retrieve the CIFAR-10-LT dataset from previous works \citep{LDAM-DRW, M2m}, re-implement WideResNet \citep{WideResNet}, and the overall training pipeline.

\subsubsection*{Communication with original authors}

We contacted the authors for clarifications on the implementation details of the algorithm. Prior works had many important implementation details such as linear learning rate warmup or deferred oversampling, so we confirmed with the authors on whether these methods were used.
\section{Introduction}

Real-world datasets often have long-tailed label distributions, for example because some classes are more rare in the real world, the acquisition source is innately biased towards a few labels, or because some classes are easier to label than others. Deep neural networks often perform poorly on less-represented classes as the model is easily biased towards majority classes and results in poor generalization for minority classes.

Well-known approaches to mitigating the class imbalance problem are re-weighting the loss and re-sampling during training. However, both approaches can encourage the model to overfit to the minority class \citep{PureNoise, M2m, Buda}.

\citet{PureNoise} proposed \textbf{O}versampling with \textbf{P}ur\textbf{e} \textbf{N}oise Images (OPeN), a new re-sampling technique of replacing some oversampled images with pure noise images. During training, OPeN replaces some images in a mini-batch with pure noise images generated at the beginning of each epoch. The probability of replacing an image $x$ of class $i$ with a noise image $x_{\text{noise}}$ is proportional to the rate of oversampling $\delta$:

\begin{equation}
\mathbb{P}(\text{Replace }x\text{ with }x_{\text{noise}} \,|\, \text{Class}=i) = \left(1 - \frac{n_i}{\max_j n_j}\right) \cdot \delta
\end{equation}

where $n_i$ is the number of samples for each class $i$.

OPeN creates mini-batches containing images from two different distributions: the CIFAR-10 distribution and the pure noise distribution. Since batch normalization (BN) \citep{BatchNorm} intrinsically assumes that the input comes from a single distribution, \citet{PureNoise} also propose Distribution Aware Routing Batch Normalization (DAR-BN) that replaces the BN layers. DAR-BN separates the pure noise images to normalize the activation maps separately from the natural images.

\section{Scope of reproducibility}
\label{sec:claims}

We investigate the following claims from \citet{PureNoise}. We list in parentheses the figures in the original paper that correspond to each claim.

\begin{enumerate}
    \item OPeN improves model performance on CIFAR-10/100-LT by improving accuracies on classes with lower frequencies. (Table 1, Figure 7)
    \item DAR-BN improves the performance of OPeN on CIFAR-10/100-LT compared to baseline Batch Normalization methods. (Table 4)
    \item The performance improvement of OPeN is robust under various data augmentation methods. (Figure 3)
    \item OPeN improves performance on the full CIFAR-10/100 dataset. (Section 5)
\end{enumerate}

\section{Methodology}

\subsection{Model descriptions}

For CIFAR-10-LT and CIFAR-100-LT datasets, \citet{PureNoise} use the WideResNet\nobreakdash-28\nobreakdash-10 \citep{WideResNet} architecture. Because the author's code was not public, we modified the implementation by \citet{torchdistill} by replacing batch normalization layers with DAR-BN.

\subsection{Datasets}

\citet{PureNoise} used 5 datasets: CIFAR-10-LT, CIFAR-100-LT, CelebA-5, ImageNet-LT, and Places-LT. As the authors only reported results from CIFAR-10-LT and CIFAR-100-LT in their ablation studies, we also focus on these two datasets.

CIFAR-10-LT and CIFAR-100-LT are long-tailed variants of the CIFAR-10 and CIFAR-100 datasets respectively, proposed by \citet{ClassBalancedLoss}. These long-tailed training datasets are created by reducing the number of training samples following an exponential function $n_i \cdot \text{IR}^{i/(C-1)}$, where $C$ is the number of classes in the dataset and $i$ is a class index from $0$ to $C-1$. IR denotes the imbalance ratio of the dataset, defined as the ratio of frequencies of the largest and smallest classes. 

\begin{table}[!ht]
    \centering
    \begin{tabular}{c|cc}
        Dataset & Imbalance Ratio (IR) & Number of training examples \\ \hline
        CIFAR-10-LT & 100 & 12406 \\
        CIFAR-10-LT & 50 & 13996 \\
        CIFAR-100-LT & 100 & 10847 \\
        CIFAR-100-LT & 50 & 12608 \\
    \end{tabular}
    \caption{Different long-tail variants of the CIFAR-10/100 datasets. A higher imbalance ratio signifies that the dataset is more imbalanced.}
    \label{tab:longtail_datasets}
\end{table}

For evaluation, we compute the accuracy using the original CIFAR-10/100 validation dataset of 10000 images. This allows for evaluation on a balanced set of examples, penalizing models that focus on majority classes during training.

For normalizing the input images, we used the per-channel mean of $(0.4914, 0.4822, 0.4465)$ and standard deviation of $(0.2023, 0.1994, 0.2010)$ for both datasets, following \citet{MiSLAS, LDAM-DRW}. However, we found them to differ from the values we computed, so we conduct additional experiments in Section~\ref{sec:input_norm_values} of the Appendix.

\subsection{Hyperparameters}
\label{sec:hyperparameters}

To provide a complete overview of the experiments, we use this section to list all the hyperparameters. For all experiments in the paper, unless specified, the experiment settings match that of Table~\ref{tab:hyperparameters}.

\begin{table}[!ht]
    \centering
    \begin{subtable}[b]{0.49\textwidth}
        \centering
        \begin{tabular}{c|c}
            Hyperparameters & Values \\
            \hline
            Model & WideResNet-28-10 \\
            Dropout rate & $0.3$ \\
            Batch size$^{*}$ & $128$ \\
            Optimizer & SGD \\
            Momentum & $0.9$ \\
            Weight decay & $2 \times 10^{-4}$ \\
        \end{tabular}
    \end{subtable}
    \hfill
    \begin{subtable}[b]{0.49\textwidth}
        \centering
        \begin{tabular}{c|c}
            Hyperparameters & Values \\
            \hline
            Initial learning rate (lr) & $0.1$ \\
            lr decay epochs & $160, 180$ \\
            lr decay gamma & $0.01$ \\
            Linear warmup epochs$^{*}$ & $5$ \\
            OPeN noise image ratio ($\delta$) & $1/3$ \\
            OPeN start epoch & $160$ \\
        \end{tabular}
    \end{subtable}
    \caption{Default hyperparameters used for experiments. $^{*}$ denote hyperparameters not described in \citet{PureNoise} but confirmed through email. Check Section~\ref{sec:communication} for more details.}
    \label{tab:hyperparameters}
\end{table}

\subsection{Experimental setup and code}

As the authors have not released the code yet, we re-implemented most of the code from the description of the paper while using open-source code from prior works. We imported long-tailed dataset generation from \citet{LDAM-DRW}, and the base WideResNet model from \citet{torchdistill}, which we modified to use DAR-BN. We used the code snippets from the original paper for noise image generation and parts of DAR-BN.

We used Weights and Biases \citep{pip__wandb} for tracking experiments, and OmegaConf, a subset of Hydra \citep{pip__hydra}, for configuring hyperparameters. All the code used to run experiments in this paper has been anonymized and submitted with the paper as supplementary material and available at \url{https://anonymous.4open.science/r/pure-noise-4166/}. It will be released on GitHub once the Reproducibility Challenge is finished.

\subsection{Computational requirements}

All experiments were performed on a cloud computing service using virtual machines with 12 vCPU, 62 GB RAM, and one NVIDIA RTX A5000 graphics card with 24 GB VRAM. Using the default experiment setting specified in Table~\ref{tab:hyperparameters}, Empirical Risk Minimization (training the model without oversampling or OPeN) took approximately 1 hour and 50 minutes. Using the checkpoints saved after 160 epochs, training with deferred oversampling took 22 minutes, and training with OPeN took 45 minutes.

\section{Results}
\label{sec:results}

\subsection{Results reproducing original paper}

\subsubsection{OPeN encourages generalization by improving minority-class accuracy}

To verify Claim 1, we trained the model using four different oversampling schemes: (i) Empirical Risk Minimization (ERM): training without oversampling (ii) Resampling (RS): sampling by weights inverse of class frequency (iii) Deferred Resampling (DRS): deferring RS to last phase of training (iv) OPeN: oversampling with pure noise during the same last phase of training. For CIFAR-10-LT (IR=100) dataset, we reproduced the mean validation accuracy of the DRS baseline and OPeN to within 0.6\% of the reported value, which supports Claim 1. For other datasets and IR ratios, the performance of DRS was not reported in the original paper. We measured the performance of DRS for those datasets because DRS is the fair baseline for OPeN as both methods use the same deferred resampling schedule. OPeN outperformed the baselines across all datasets.

\begin{table}[!ht]
    \centering
    \begin{tabular}{c|cc}
        Source & Reported \citep{PureNoise} & Ours \\
        \hline
        \multirow{1}*{ERM} & 79.6 & 81.18 \\
        \multirow{1}*{RS} & 75.1 & 74.82 \\
        \multirow{1}*{DRS} & 83.0 & 83.22 \\
        \multirow{1}*{OPeN} & 84.6 & 85.04 \\
    \end{tabular}
    \caption{Comparing accuracy of resampling schemes on CIFAR-10-LT (IR=$100$) dataset. Reported accuracy are from Table 1 in \citet{PureNoise}.}
    \label{tab:accuracy_comparisons}
\end{table}

We also compute the per-class accuracies to understand if the improvement is from minority classes. Indeed, we confirm that compared to DRS, OPeN improves the performance of the two least frequent classes by 8.2\% while sacrificing only 0.9\% accuracy for the two most frequent classes. For a complete comparison, we ask the readers to look at Figure~\ref{fig:group_accuracy} and Figure~\ref{fig:group_accuracy_original} in the Appendix.

\subsubsection{DAR-BN outperforms other batch normalization layers when used with OPeN}

To verify Claim 2, we trained three models with different batch normalization: (i) Standard BN \citep{BatchNorm}: normalizing pure noise and natural activation maps together using one BN layer (ii) Auxiliary BN \citep{AuxBN}: normalizing pure noise and natural activation maps separately using two BN layers (iii) Distribution-Aware Routing BN (DAR-BN) \citep{PureNoise}: using the affine parameters learned from natural activation maps to normalize noise activation maps. For CIFAR-10-LT (IR=100) and CIFAR-100-LT (IR=100) datasets, DAR-BN outperformed Standard BN and Auxiliary BN in terms of mean validation accuracy (Table~\ref{tab:batchnorm_ablation}). This supports the claim and shows that DAR-BN is essential to the success of OPeN, as without DAR-BN, the accuracy is lower than the accuracy of DRS (83.22). In Table~\ref{tab:resnet_batchnorm_ablation}, we also perform the same experiment on ResNet and come to the same conclusion, further validating the claim.

\begin{table}[!ht]
    \centering
    \begin{tabular}{c|cc|cc}
        Dataset & \multicolumn{2}{c|}{CIFAR-10-LT} & \multicolumn{2}{c}{CIFAR-100-LT} \\
        Source & Reported \citep{PureNoise} & \;\;\;\;\;Ours\;\;\;\;\; & Reported \citep{PureNoise} & \;\;\;\;\;Ours\;\;\;\;\; \\
        \hline
        Standard BN \citep{BatchNorm} & 81.45 & 81.81 & 49.18 & 49.26 \\
        Auxiliary BN \citep{AuxBN} & 83.38 & 82.23 & 50.13 & 51.27 \\
        DAR-BN \citep{PureNoise} & 84.64 & 85.04 & 51.50 & 52.12 \\
    \end{tabular}
    \caption{Ablation experiment: comparing DAR-BN with other Batch Normalization layers (IR=$100$). Reported scores are from Table 4 in \citet{PureNoise}.}
    \label{tab:batchnorm_ablation}
\end{table}

\subsubsection{OPeN is robust to various data augmentation methods} To verify Claim 3, we compared ERM, DRS, and OPeN on CIFAR-10-LT (IR=100) dataset using three data augmentations of increasing strength: (i) random horizontal flip and random 32x32 pixel crop with padding of 4 (ii) add Cutout, \citep{CutOut} which zeros out one 16x16 pixel patch (iii) add SimCLR, \citep{SimCLR} which randomly applies color jitter, grayscale, and Gaussian blur. OPeN outperformed DRS and ERM across all augmentations, which supports the claim. We forgo AutoAugment \citep{AutoAugment} for this ablation study because AutoAugment was optimized using the full balanced dataset and is an unfair augmentation strategy for the imbalanced sub-dataset \citep{PureNoise, GLICO}.

\begin{table}[!ht]
    \centering
    \begin{tabular}{c|cc|cc|cc}
         & \multicolumn{2}{c|}{Flip and Crop} & \multicolumn{2}{c|}{Add Cutout} & \multicolumn{2}{c}{Add SimCLR} \\
        Source & Reported & Ours & Reported & Ours & Reported & Ours \\
        \hline
        ERM & 74.3 & 74.6 & 77.7 & 78.7 & 79.6 & 80.7 \\
        DRS \citep{LDAM-DRW} & 76.5 & 75.4 & 80.3 & 79.5 & 83.0 & 83.2 \\
        OPeN \citep{PureNoise} & 80.3 & 79.9 & 83.1 & 83.9 & 84.6 & 84.3 \\
    \end{tabular}
    \caption{Data augmentation ablation experiment. Reported accuracy are from Figure 3 in \citet{PureNoise}.}
    \label{tab:dataaug_ablation}
\end{table}

\subsubsection{Adding pure noise improves performance on balanced datasets}

In Claim 4, the authors propose that using pure noise is useful as a general data augmentation method beyond imbalanced datasets. That is, given a balanced dataset, we can simply add a fixed number of pure noise images to each class and train with DAR-BN. Since this approach does not modify natural images, it can complement any existing data augmentations. The authors experiment by adding random noise images with a fixed noise-to-natural ratio of $1:4$ in each batch and report percentage improvement over training without random noise images. The authors used different hyperparameters, such as Adam optimizer with $\beta_1 = 0.9$, $\beta_2 = 0.999$ and AutoAugment \citep{AutoAugment} for data augmentation. Furthermore, we communicated with the authors to find that a fixed learning rate of $0.001$ was used without linear warmup \citep{ImageNet1h}. Our results showed that adding pure noise images improve the performance on the balanced dataset.

\begin{table}[!ht]
    \centering
    \begin{tabular}{c|c|c|c}
        Source & Improvement & Baseline Accuracy & Pure Noise Accuracy \\ \hline
        Reported & +0.9\% & - & - \\ 
        Ours & +1.6\% & 87.16 & 88.57 \\
    \end{tabular}
    \caption{Performance improvement of OPeN on the full balanced CIFAR-10 dataset. The original paper reported the percentage improvement but not the baseline and pure noise accuracy.}
    \label{tab:balanced_results}
\end{table}

\subsection{Results beyond original paper}

\subsubsection{ResNet architecture}

\citet{PureNoise} used WideResNet-28-10 for their experiments with CIFAR-10 and CIFAR-100. However, prior works \citep{LDAM-DRW, M2m, MiSLAS} used a smaller ResNet-32 network. To compare performance with results originally reported by prior works, we train OPeN on the ResNet architecture. For these experiments, we used the ResNet-32 implementation by \citet{ResNet_akamaster} and replaced batch normalization layers with DAR-BN.

In Table~\ref{tab:resnet32_results}, we compare OPeN with the performance reported by prior works. We find that OPeN still shows improvement over ERM, RS, and DRS. However, the comparative advantage of OPeN compared to LDAM-DRW or M2m is less apparent in ResNet-32 with Flip and Crop augmentation, compared to the authors' original result with WideResNet-28-10 with SimCLR augmentation. This suggests that the performance improvement from OPeN may be more orthogonal to the improvement caused by a bigger network or more complex data augmentation.

\begin{table}[!ht]
    \centering
    \begin{subtable}[t]{0.49\textwidth}
        \centering
        \begin{tabular}[t]{c|c|c}
            Source & Method & Accuracy \\
            \hline
            \multirow{2}*{LDAM-DRW \citep{LDAM-DRW}} & ERM & $70.36$ \\
                                                     & LDAM-DRW & $77.03$ \\
            \hline
            \multirow{4}*{M2m \citep{M2m}} & ERM & $68.7 \pm 1.43$ \\
                                           & RS & $70.4 \pm 1.15$ \\
                                           & DRS & $75.2 \pm 0.26$ \\
                                           & M2m & $78.3 \pm 0.16$ \\
        \end{tabular}
    \end{subtable}
    \hfill
    \begin{subtable}[t]{0.49\textwidth}
        \centering
        \begin{tabular}[t]{c|c|c}
            Source & Method & Accuracy \\
            \hline
            \multirow{4}*{Ours} & ERM & $71.70$ \\
                                & RS & $70.09$ \\
                                & DRS & $75.78$ \\
                                & OPeN & $77.52$ \\
        \end{tabular}
        \end{subtable}
    \caption{Performance of ResNet-32 models for CIFAR-10-LT (IR=$100$). Note that for M2m \citep{M2m}, a different dataset variant has been used. Check Section~\ref{sec:dataset_random_seed} for more information.}
    \label{tab:resnet32_results}
\end{table}

We also perform ablation studies on the effect of DAR-BN on the ResNet architecture and find that DAR-BN improves performance, supporting the central claim by the authors (Table~\ref{tab:batchnorm_ablation}).

\begin{table}[!ht]
    \centering
    \begin{tabular}{c|c}
        BN Layer & Accuracy \\
        \hline
        Standard BN \citep{BatchNorm} & 74.37 \\
        Auxiliary BN \citep{AuxBN} & 75.08 \\
        DAR-BN \citep{PureNoise} & 77.52 \\
    \end{tabular}
    \caption{Batch normalization ablation experiment for OPeN. Same experiment setting as Table~\ref{tab:batchnorm_ablation}, but with ResNet-32 and Flip and Crop augmentation.}
    \label{tab:resnet_batchnorm_ablation}
\end{table}

Finally, we experimented with the ResNet-32 network on a full, balanced CIFAR-10 dataset. Unlike when using WideResNet (Table~\ref{tab:balanced_results}), adding pure noise showed slightly lower performance.

\begin{table}[!ht]
    \centering
    \begin{tabular}{c|c|c}
        Accuracy without pure noise & Accuracy with pure noise & Change \\ \hline
        86.51 & 86.19 & -0.37\% \\
    \end{tabular}
    \caption{Performance of adding pure noise to the full balanced CIFAR-10 dataset when using ResNet-32 model.}
    \label{tab:balanced_results_resnet32}
\end{table}

\subsubsection{Random seed for long-tailed dataset generation}
\label{sec:dataset_random_seed}

The CIFAR-10-LT dataset is a subset of the CIFAR-10 dataset, so different random seed creates a different dataset. This can be problematic as different papers use different training data, resulting in an unfair comparison of methods.

\citet{ClassBalancedLoss} did not set a random seed but saved their datasets in tfrecords format. Later works implemented their own version of the long-tailed dataset and set a random seed. We compared the downloaded images from \citet{ClassBalancedLoss} and ran the dataset generation code from \citet{LDAM-DRW} and \citet{M2m} and discovered that the resulting CIFAR-10-LT datasets have a considerable amount of different training images. On average, each training dataset has around 25\% unique images.\footnote{We refer the readers to the Appendix for a visualization the intersection of datasets (Figure~\ref{fig:cifar10_indices_venn}) and for an example of unique images for each dataset (Figure~\ref{fig:cifar10_indices_differences}).}

To analyze the effect of this discrepancy in the training dataset, we trained the model on each variant. We find that the long-tail subset used to train the model results in a noticeable change in performance. For our work, we use the subset by \citet{LDAM-DRW}, which gave accuracy scores closest to that reported by \citet{PureNoise}. We ask future researchers to specify the long-tail subset they used for reproducibility, and we list the indices of images used for each variant in our code.

\begin{table}[!ht]
    \centering
    \begin{tabular}{c|ccc}
        Source & ERM & DRS & OPeN \\ \hline
        \citet{ClassBalancedLoss} & 79.26 & 80.87 & 84.19 \\
        \citet{M2m} & 78.37 & 81.64 & 87.11 \\
        \citet{LDAM-DRW} & 81.18 & 83.22 & 85.04 \\
        \hline
        Reported by \citet{PureNoise} & 79.6 & 83.0 & 84.6 \\
    \end{tabular}
    \caption{Performance of models trained on different CIFAR-10-LT (IR=100) datasets from various sources.}
    \label{tab:cifar10_indices_perf}
\end{table}

\subsubsection{Analysis of model priors}

\citet{PureNoise} hypothesized that the enhanced performance of OPeN may be due to the shift in model priors. We perform experiments to understand to which degree OPeN influences the model prior. We test three hypotheses:

\begin{enumerate}
    \item OPeN encourages the model to encode noise and out-of-distribution images similar to minority images.
    \item OPeN results in noise and out-of-distribution images being classified as minority images.
    \item Model trained with OPeN predicts any image as a minority class more often.
\end{enumerate}

\begin{figure}[!ht]
    \centering
    \begin{subfigure}[b]{0.49\textwidth}
        \centering
        \includegraphics[width=\textwidth]{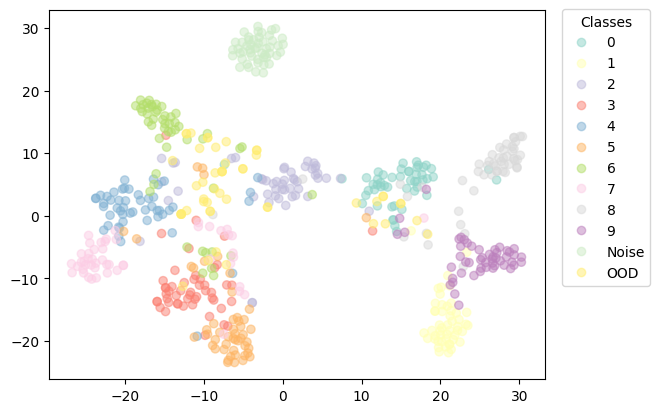}
        \caption{Each class colored differently}
        \label{fig:tsne_open}
    \end{subfigure}
    \hfill
    \begin{subfigure}[b]{0.49\textwidth}
        \centering
        \includegraphics[width=\textwidth]{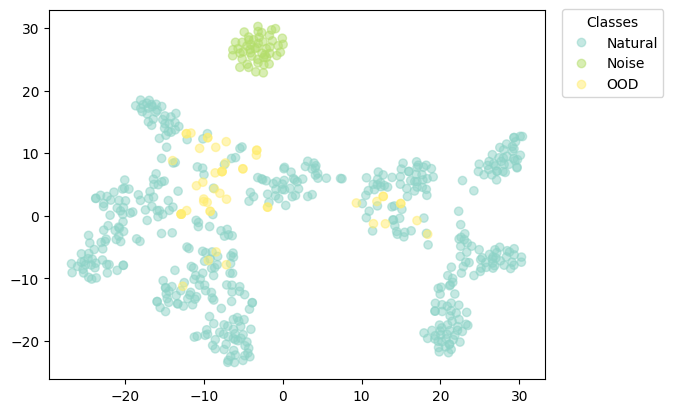}
        \caption{Only noise and OOD images colored differently}
        \label{fig:tsne_ternary_open}
    \end{subfigure}
    \caption{t-SNE of 50 examples from each class in the CIFAR-10 validation dataset, 50 out-of-distribution images from the CIFAR-100 validation dataset, and 50 random pure noise. Images are embedded with a model trained with OPeN. Noise images are gathered into one cluster that is separable from any other class clusters, whereas OOD images are more dispersed across multiple classes. This phenomenon is not unique to OPeN, as it also appears with DRS (Figure~\ref{fig:tsne_parent_drs}).}
    \label{fig:tsne_parent}
\end{figure}

Following \citet{M2m}, we use t-SNE \citep{tsne} to visualize the embeddings generated by the network. Embeddings are computed from 50 randomly chosen samples from the validation set of CIFAR-10 using the features from the penultimate layer of the WideResNet network. To represent out-of-distribution (OOD) images, we sample 50 images from one class of the CIFAR-100 validation dataset, as its classes are mutually exclusive to CIFAR-10 \cite{TinyImages}. For noise images, we generate 50 new pure noise images. t-SNE is used on these embeddings and is visualized in Figure~\ref{fig:tsne_parent}. For both noise and OOD images, we do not see any noticeable proximity to any of the minority classes.

For the second hypothesis, we generate 1000 pure noise images and sample 1000 out-of-distribution images across all 100 classes from CIFAR-100 and pass them through a trained model. We compare the predictions of the ERM, DRS, and the OPeN model. The model trained with OPeN is less likely to predict OOD images as a majority class and more likely to be predicted as a minority class, confirming our hypothesis (Figures~\ref{fig:noise_image_classification} and \ref{fig:ood_image_classification} in Appendix).

We note that all pure noise images are classified as classes 2, 4, or 6, whereas the OOD images are more dispersed. As seen in Figure~\ref{fig:tsne_parent}, pure noise images are more clustered together, resulting in predictions gathered in a few classes, whereas for out-of-distribution images not seen during training, the predictions are more evenly distributed.

\begin{figure}[!ht]
    \centering
    \begin{subfigure}[b]{0.3\textwidth}
        \centering
        \includegraphics[width=\textwidth]{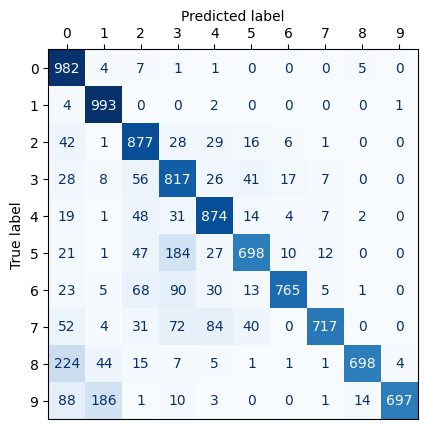}
        \caption{ERM}
        \label{fig:cmatrix_erm}
    \end{subfigure}
    \hfill
    \begin{subfigure}[b]{0.3\textwidth}
        \centering
        \includegraphics[width=\textwidth]{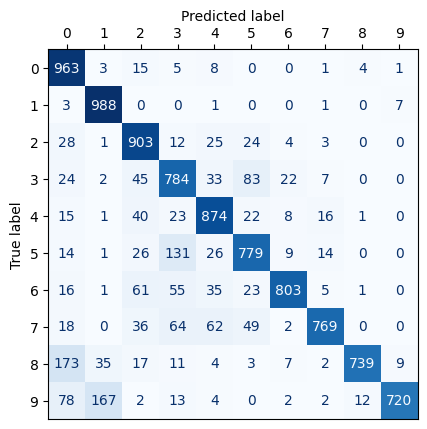}
        \caption{Deferred oversampling}
        \label{fig:cmatrix_drs}
    \end{subfigure}
    \hfill
    \begin{subfigure}[b]{0.3\textwidth}
        \centering
        \includegraphics[width=\textwidth]{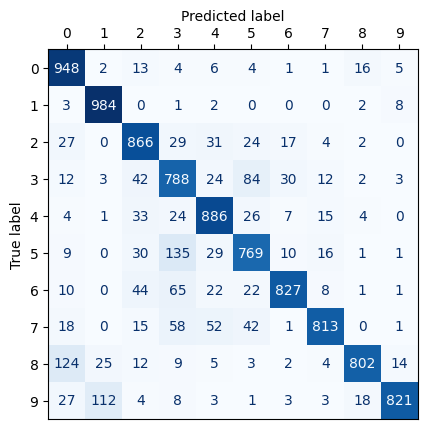}
        \caption{OPeN}
        \label{fig:cmatrix_open}
    \end{subfigure}
    \caption{Confusion matrices of models trained on CIFAR-10-LT (IR=100).}
    \label{fig:confusion_matrices}
\end{figure}

To verify the final hypothesis, we plot the confusion matrix of models trained with ERM, DRS, and OPeN in Figure~\ref{fig:confusion_matrices}. We find that models trained in OPeN are less likely to have images in the minority classes predicted as one of the majority classes.

To conclude, we find that training the model with OPeN makes the model more likely to classify any input image as a minority class. However, this is not done by embedding the noise images to be similar to the minority classes.

\section{Discussion}
Our experiments support the four claims by \citet{PureNoise}. First, our results showed that OPeN improves the mean test accuracy over DRS and other baseline resampling methods across CIFAR-10-LT (IR=100, 50) and CIFAR-100-LT (IR=100, 50) datasets. We confirmed that this improvement is driven by a significant improvement in the accuracy of minority classes. Also, our ablation study supports the claim that using the affine parameters learned from natural activation to normalize the noise activations (DAR-BN) is crucial to the performance of OPeN. Moreover, our experiments showed that OPeN is robust to various data augmentation methods, as OPeN outperforms baseline resampling methods across data augmentations of varying strengths. Finally, our results showed that adding pure noise can be used as an additional data augmentation method to improve the performance on a full, balanced CIFAR-10 dataset.

Then, we ran experiments using a smaller ResNet-32 network and Flip and Crop augmentations to compare with the performance reported by prior works. OPeN still showed improvement over ERM, RS, and DRS, and DAR-BN showed improvement over Standard and Auxiliary BN. However, the comparative advantage of OPeN to preceding papers was less apparent when using a smaller model and simpler data augmentations, which suggests that performance improvement from OPeN may be more orthogonal to the improvement caused by a bigger network or more complex data augmentation. Also, we noticed that adding pure noise to the balanced CIFAR-10 dataset slightly lowered the performance when using ResNet-32.

Beyond the original paper, we proposed three hypotheses to understand if the enhanced performance of OPeN is due to a shift in model priors. Our analysis shows that OPeN makes the model more likely to classify pure noise, OOD, and misclassified test images as a minority class. However, our visualization suggests that this is not done by encoding the pure noise or OOD images to be similar to the images from minority classes.

Furthermore, our investigation into the preceding papers in imbalanced classification suggests directions to improve the reproducibility of future work in this domain. We found that the images in two instances of the CIFAR-10-LT dataset can vary significantly depending on the random seed used for sampling from the full, balanced CIFAR-10 dataset. Also, prior work used varying mean and standard deviation, which are sometimes computed from the full balanced dataset, for input normalization (Tables~\ref{tab:cifar10_input_norm_values} and \ref{tab:cifar100_input_norm_values}). Hence, more detailed documentation for generating the long-tailed dataset and computing the dataset statistics for input normalization may help improve reproducibility and fair comparison across papers.

\subsection{What was easy and what was difficult}
The authors provided two functions that (i) given a batch, samples noise indices and replaces corresponding natural images with pure noise (ii) given a batch, noise indices, and a BN layer, applies DAR-BN. These functions were clearly documented with docstrings and were easy to use. The authors also clearly documented the key hyperparameters that were used in the main experiments.

Aside from the core functions, the authors' code was not available, so we had to fully implement it based on the description of the paper. Hence, reproducing the reported performance on the first dataset took more time than we initially anticipated, as we had to study available code from previous related papers, including \citet{M2m} and \citet{LDAM-DRW}. For example, the paper described clipping the sampled noise images to $[0, 1]$, but we did not find a corresponding operation in the provided functions. We found that the InputNormalize module from \citet{M2m} had clipping already implemented, so we imported the module and fit it into our training workflow. Following \citet{LDAM-DRW}, the authors used two different resampling baselines: one baseline started oversampling from the first epoch (Table 1 of \citep{PureNoise}), and another baseline started oversampling from the 160th epoch (Figure 3 of \citep{PureNoise}). The difference between these two baselines was unclear and needed clarification from the authors. Also, we verified a few available implementations of WideResNet-28-10 \citep{WideResNet} and ResNet-32 \citep{ResNet} for correctness. 

Yet, comparing prior related works revealed interesting discrepancies as well, such as the differences in the generated long-tailed dataset and input normalization.

\subsection{Communication with original authors}
\label{sec:communication}

Overall, we found the paper to be reproducible, as we were able to validate the effectiveness of OPeN before contacting the authors for clarification. However, we struggled to match the performance of baseline methods. We were able to contact the authors through email to confirm the following details:

\begin{itemize}[noitemsep]
    \item A batch size of 128 was used during training.
    \item Learning rate warm-up \citep{ImageNet1h} was used for the first 5 epochs.
    \item Effective number of samples \citep{ClassBalancedLoss} was not used for calculating oversampling weights.
    \item Oversampling did not increase the number of examples seen per epoch.
    \item A fixed learning rate of 0.001 was used for training on the full CIFAR-10 dataset.
    \item A dropout rate of 0.3 was used for WideResNet.
\end{itemize}
 
\hypersetup{linkcolor=black,urlcolor=darkgray}
\renewcommand\emph[1]{{\bfseries #1}}
\setlength\bibitemsep{0pt}
\printbibliography

\newpage
\appendix

\section{Additional results beyond original paper}
\subsection{Performance of OPeN on additional dataset and imbalance ratios}

In addition to CIFAR-10-LT (IR=100) dataset, we compared the performance of OPeN \citep{PureNoise} to DRS \citep{LDAM-DRW}, RS, and ERM on CIFAR-10-LT (IR=50) and CIFAR-100-LT (IR=100,50) datasets. Consistent with claim 1, OPeN outperformed the baseline resampling schemes across all datasets. The original paper did not report the accuracy of deferred resampling (DRS) \citep{LDAM-DRW} for these additional datasets. Nonetheless, we compared OPeN with DRS because DRS provides a fair baseline, as OPeN uses the same deferred resampling schedule.

\begin{table}[!ht]
    \centering
    \begin{tabular}{c|cc|cc|cc}
        Dataset & \multicolumn{2}{c|}{CIFAR-10-LT} & \multicolumn{4}{c}{CIFAR-100-LT} \\
        IR & \multicolumn{2}{c|}{50} & \multicolumn{2}{c}{100} & \multicolumn{2}{c}{50} \\
         & Reported \citep{PureNoise} & Ours & Reported \citep{PureNoise} & Ours & Reported \citep{PureNoise} & Ours \\
        \hline
        \multirow{1}*{ERM} & 84.9 & 84.9 & 47.0 & 47.1 & 52.4 & 52.7 \\
        \multirow{1}*{RS} & 82.2 & 80.9 & 42.5 & 41.6 & 48.0 & 46.5 \\
        \multirow{1}*{DRS} & - & 86.9 & - & 50.8 & - & 55.8 \\
        \multirow{1}*{OPeN} & 87.9 & 87.8 & 51.5 & 52.1 & 56.3 & 56.5 \\
    \end{tabular}
    \caption{Comparison of accuracy on CIFAR-10-LT (IR=50) and CIFAR-100-LT (IR=100,50).}
    \label{tab:accuracy_comparisons_extra}
\end{table}

\subsection{Hyperparameter Search}

\subsubsection{Input normalization values}
\label{sec:input_norm_values}

The authors did not specify the mean and standard deviation used to normalize the dataset. We explored various prior works \citep{LDAM-DRW, M2m, MiSLAS} and discovered that they differed from the values we computed (Tables~\ref{tab:cifar10_input_norm_values}~and~\ref{tab:cifar100_input_norm_values}). Surprisingly, we found that many prior works use mean values from the full CIFAR-10/100 datasets instead of the values from the long-tailed variants. This could result in an unfair evaluation, as the statistics from the full training dataset may resemble the validation dataset, as they are both balanced.

\begin{table}[!ht]
    \centering
    \begin{tabular}{c|cc}
        Source & Mean & Std \\ \hline
        \citet{MiSLAS} & (0.4914, 0.4822, 0.4465) & (0.2023, 0.1994, 0.2010) \\
        \citet{LDAM-DRW} & (0.4914, 0.4822, 0.4465) & (0.2023, 0.1994, 0.2010) \\
        \citet{M2m} & (0.4914, 0.4822, 0.4465) & (0.2023, 0.1994, 0.2010) \\ \hline
        CIFAR-10$^{\dagger}$ & (0.4914, 0.4822, 0.4465) & (0.2470, 0.2435, 0.2616) \\ 
        CIFAR-10-LT (IR=50)$^{\dagger}$ & (0.4978, 0.5003, 0.4840) & (0.2505, 0.2477, 0.2722) \\
        CIFAR-10-LT (IR=100)$^{\dagger}$ & (0.4989, 0.5044, 0.4926) & (0.2513, 0.2485, 0.2734) \\
    \end{tabular}
    \caption{Per-channel mean and standard deviations for experiments on CIFAR-10-LT. Values calculated in this paper are marked by $\dagger$.}
    \label{tab:cifar10_input_norm_values}
\end{table}

\begin{table}[!ht]
    \centering
    \begin{tabular}{c|cc}
        Source & Mean & Std \\ \hline
        \citet{MiSLAS} & (0.4914, 0.4822, 0.4465) & (0.2023, 0.1994, 0.2010) \\
        \citet{LDAM-DRW} & (0.4914, 0.4822, 0.4465) & (0.2023, 0.1994, 0.2010) \\
        \citet{M2m} & (0.5071, 0.4867, 0.4408) & (0.2675, 0.2565, 0.2761) \\ \hline
        CIFAR-100$^{\dagger}$ & (0.5071, 0.4866, 0.4409) & (0.2673, 0.2564, 0.2762) \\
        CIFAR-100-LT (IR=50)$^{\dagger}$ & (0.5202, 0.4916, 0.4415) & (0.2676, 0.2609, 0.2778) \\
        CIFAR-100-LT (IR=100)$^{\dagger}$ & (0.5228, 0.4929, 0.4420) & (0.2677, 0.2617, 0.2780) \\
    \end{tabular}
    \caption{Per-channel mean and standard deviations for experiments on CIFAR-100-LT. Values calculated in this paper are marked by $\dagger$.}
    \label{tab:cifar100_input_norm_values}
\end{table}

We train a WideResNet network with the default experiment setting in Section~\ref{sec:hyperparameters} to investigate the effect of these values. As shown in Table~\ref{tab:cifar10_input_norm_perf}, using the calculated mean and standard deviation from the long-tailed dataset reduced performance. Regardless, we do find that OPeN still improves performance over ERM and DRS, supporting the central claim of \citet{PureNoise}.

\begin{table}[!ht]
    \centering
    \begin{tabular}{c|ccc}
        Source & ERM & DRS & OPeN \\ \hline
        Baseline \citep{MiSLAS, LDAM-DRW, M2m} & 81.18 & 83.22 & 85.04 \\
        CIFAR-10$^{\dagger}$ & 80.50 & 82.39 & 84.72 \\ 
        CIFAR-10-LT (IR=100)$^{\dagger}$ & 79.28 & 81.03 & 84.12 \\
    \end{tabular}
    \caption{Performance on different input normalization values on CIFAR-10-LT (IR=100). Values calculated in this paper are marked by $\dagger$.}
    \label{tab:cifar10_input_norm_perf}
\end{table}

\subsubsection{Batch size}

Before we communicated with the authors and confirmed that the batch size used was 128, we performed a hyperparameter search ourselves. In Table~\ref{tab:batch_size}, we list the batch sizes and their respective performance. Experiments show that 128 is the best batch size for the set of hyperparameters.

\begin{table}[!ht]
    \centering
    \begin{tabular}{c|ccc}
        \multirow{2}{*}{Batch size} & \multicolumn{3}{c}{Accuracy} \\
         & ERM & DRS & OPeN \\ \hline
        32 & 74.38 & 80.15 & 80.50 \\
        64 & 78.69 & 82.89 & 83.89 \\
        128 & \textbf{81.18} & \textbf{83.22} & \textbf{85.04} \\
        256 & 79.11 & 75.28 & 82.36 \\
        512 & 75.19 & 71.39 & 79.22 \\
    \end{tabular}
    \caption{Hyperparameter search for batch size. Experiment was done on CIFAR-10-LT (IR=100) with the default setting in Table~\ref{tab:hyperparameters} except for the batch size. Results with highest accuracy for each method are boldfaced.}
    \label{tab:batch_size}
\end{table}

\subsubsection{Noise ratio}

The noise ratio is the hyperparameter that defines the probability of replacing an oversampled image with pure noise. The authors used a noise ratio of $\frac{1}{3}$ across all datasets. We compared the performance of OPeN across increasing levels of noise ratios. Surprisingly, replacing up to $\frac{2}{3}$ of the oversampled images with pure noise continued to provide higher validation accuracy than DRS, while the train accuracy dropped with increasing noise ratio, as expected.

\begin{table}[!ht]
    \centering
    \begin{tabular}{c|c|c|c|c|c|c}
        Noise ratio & $\sfrac{1}{6}$ & \textbf{$\sfrac{2}{6}$} & $\sfrac{3}{6}$ & $\sfrac{4}{6}$ & $\sfrac{5}{6}$ & $\sfrac{6}{6}$ \\
        \hline
        Accuracy & 83.23 & \textbf{85.04} & 84.34 & 84.23 & 83.31 & 80.01
    \end{tabular}
    \caption{Hyperparameter search for noise ratio. Experiment was done on CIFAR-10-LT (IR=100) with the default setting in Table~\ref{tab:hyperparameters} except for the noise ratio. Result with highest accuracy is boldfaced.}
    \label{tab:noise_ratio}
\end{table}

\newpage
\section{Additional Figures}

\begin{figure}[!ht]
    \centering
    \includegraphics[width=0.8\textwidth]{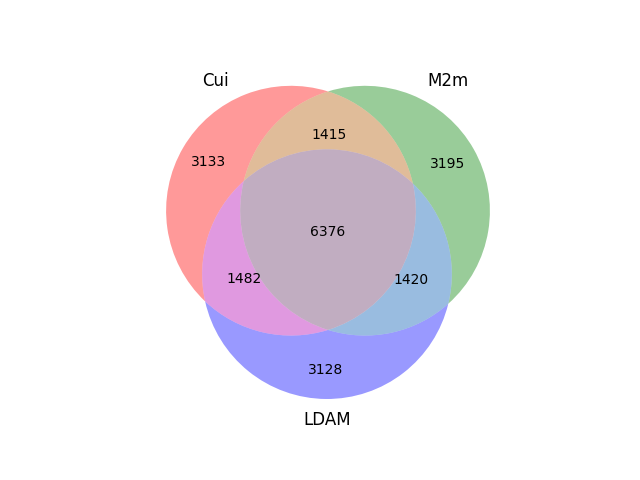}
    \caption{Venn diagram showing the intersections of training dataset used by \citet{ClassBalancedLoss} (denoted \textbf{Cui}), \citet{LDAM-DRW} (\textbf{LDAM}), and \citet{M2m} (\textbf{M2m}) for CIFAR-10-LT (IR=100).}
    \label{fig:cifar10_indices_venn}
\end{figure}

\begin{figure}[!ht]
    \centering
    \begin{subfigure}[b]{0.3\textwidth}
        \centering
        \includegraphics[width=\textwidth]{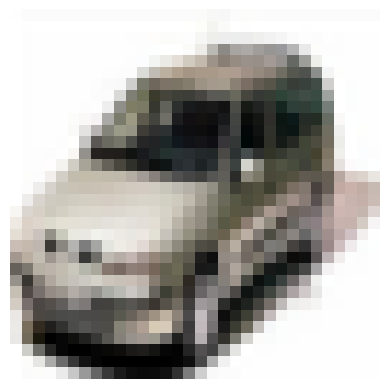}
        \caption{Image used only by \citet{ClassBalancedLoss} for training}
        \label{fig:cui_unique}
    \end{subfigure}
    \hfill
    \begin{subfigure}[b]{0.3\textwidth}
        \centering
        \includegraphics[width=\textwidth]{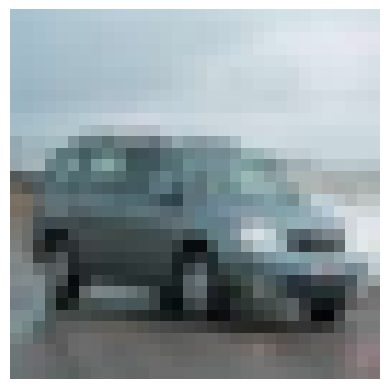}
        \caption{Image used only by \citet{M2m} for training}
        \label{fig:m2m_unique}
    \end{subfigure}
    \hfill
    \begin{subfigure}[b]{0.3\textwidth}
        \centering
        \includegraphics[width=\textwidth]{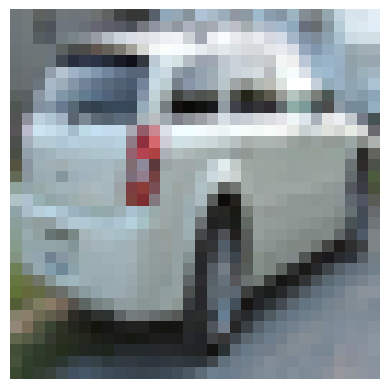}
        \caption{Image used only by \citet{LDAM-DRW} for training}
        \label{fig:ldam_unique}
    \end{subfigure}
    \caption{Examples of automobile images only found in one but not the other 2 CIFAR-10-LT (IR=100) datasets.}
    \label{fig:cifar10_indices_differences}
\end{figure}

\begin{figure}[!ht]
    \centering
    \begin{subfigure}[b]{0.49\textwidth}
        \centering
        \includegraphics[width=\textwidth]{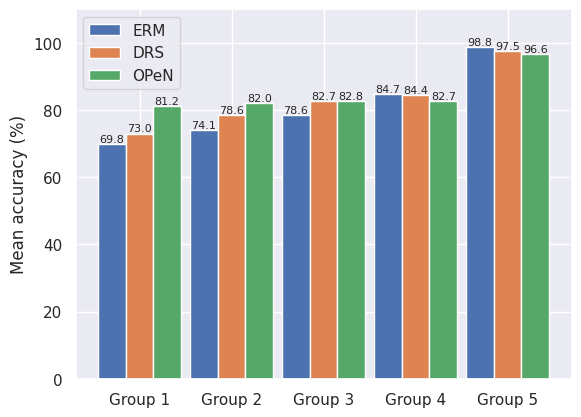}
        \caption{CIFAR-10-LT (IR=100)}
        \label{fig:group_accuracy_cifar10}
    \end{subfigure}
    \hfill
    \begin{subfigure}[b]{0.49\textwidth}
        \centering
        \includegraphics[width=\textwidth]{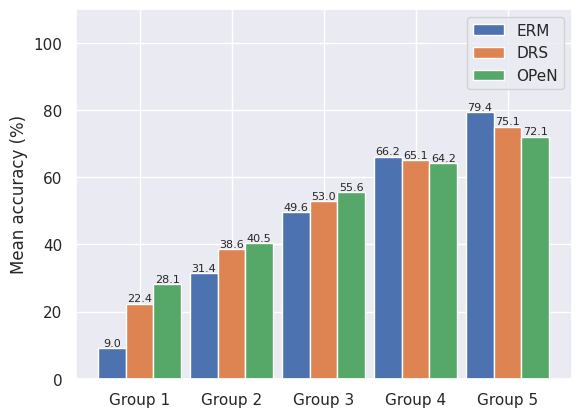}
        \caption{CIFAR-100-LT (IR=100)}
        \label{fig:group_accuracy_cifar100}
    \end{subfigure}
    \caption{Validation accuracy for CIFAR-10-LT and CIFAR-100-LT with IR=100. Classes partitioned into 5 groups, where Group 1 is the least frequent and Group 5 is the most frequent. Reproduction of Figure~\ref{fig:group_accuracy_original}.}
    \label{fig:group_accuracy}
\end{figure}

\begin{figure}[!ht]
    \centering
    \includegraphics[width=\textwidth]{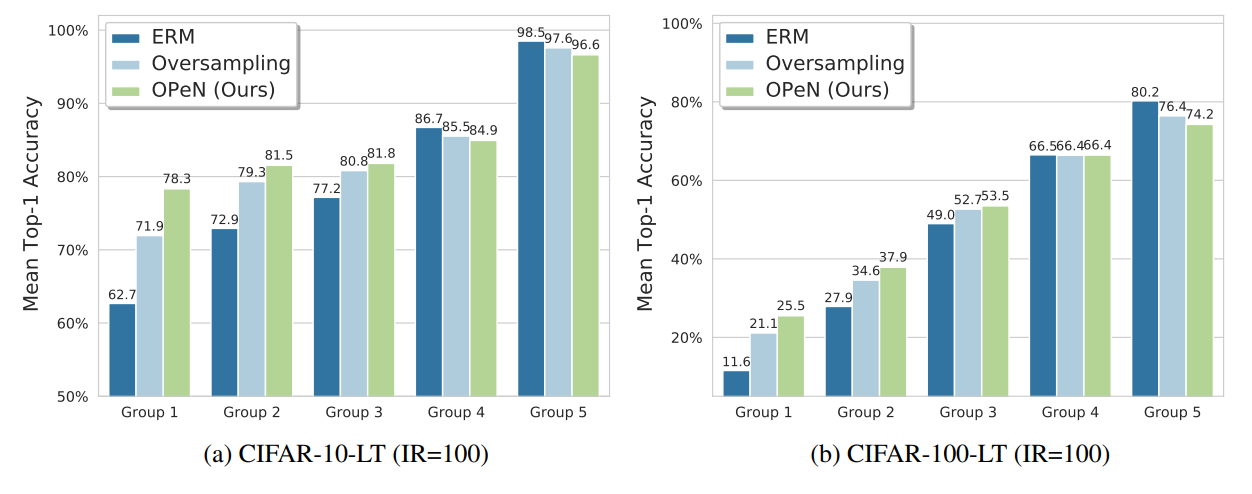}
    \caption{Figure from \citet{PureNoise} reporting validation accuracy for CIFAR-10-LT and CIFAR-100-LT with IR=100 where classes are partitioned into 5 groups. Group 1 is the least frequent and Group 5 is the most frequent.}
    \label{fig:group_accuracy_original}
\end{figure}

\begin{figure}[!ht]
    \centering
    \begin{subfigure}[b]{0.49\textwidth}
        \centering
        \includegraphics[width=\textwidth]{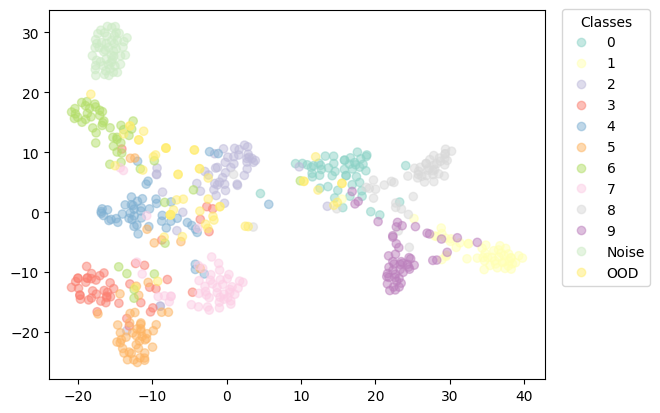}
        \caption{Each class colored differently}
        \label{fig:tsne_drs}
    \end{subfigure}
    \hfill
    \begin{subfigure}[b]{0.49\textwidth}
        \centering
        \includegraphics[width=\textwidth]{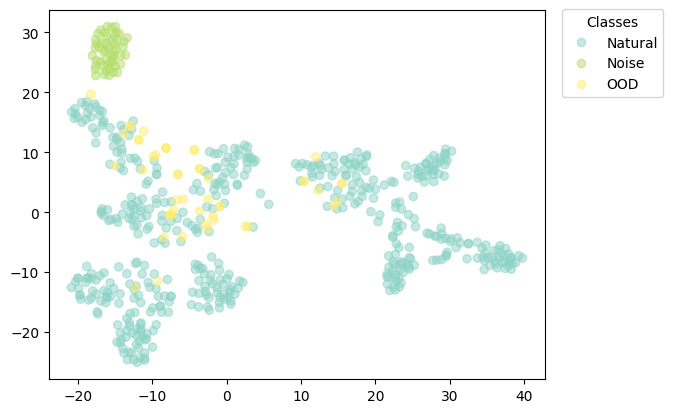}
        \caption{Only noise and OOD images colored differently}
        \label{fig:tsne_ternary_drs}
    \end{subfigure}
    \caption{t-SNE of 50 examples from each class in the CIFAR-10 validation dataset, 50 out-of-distribution images from CIFAR-100 validation dataset, and 50 random pure noise. Images are embedded with model trained with DRS.}
    \label{fig:tsne_parent_drs}
\end{figure}

\begin{figure}[!ht]
    \centering
    \begin{subfigure}[b]{0.32\textwidth}
        \centering
        \includegraphics[width=1.1\textwidth]{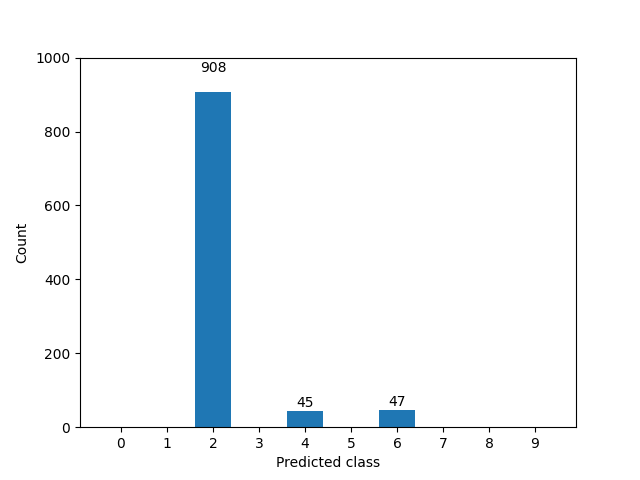}
        \caption{ERM}
        \label{fig:noise_image_classification_erm}
    \end{subfigure}
    \hfill
    \begin{subfigure}[b]{0.32\textwidth}
        \centering
        \includegraphics[width=1.1\textwidth]{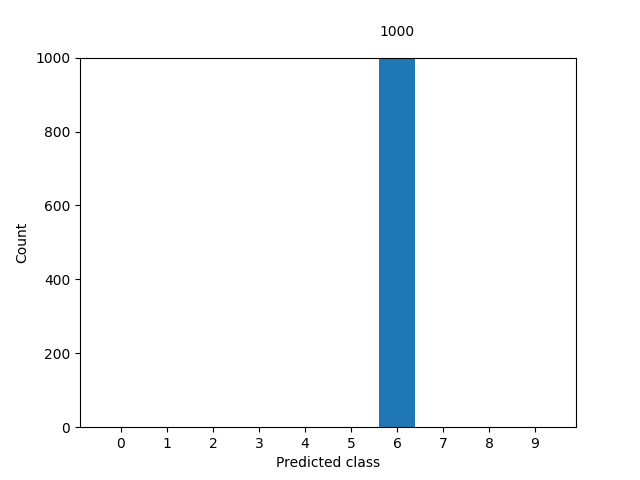}
        \caption{DRS}
        \label{fig:noise_image_classification_drs}
    \end{subfigure}
    \hfill
    \begin{subfigure}[b]{0.32\textwidth}
        \centering
        \includegraphics[width=1.1\textwidth]{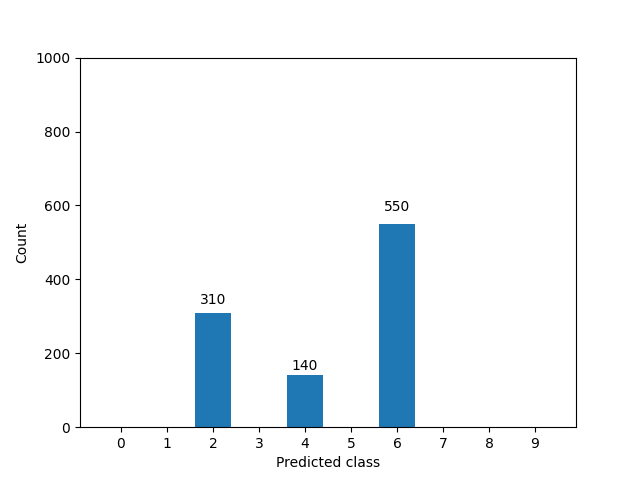}
        \caption{OPeN}
        \label{fig:noise_image_classification_open}
    \end{subfigure}
    \caption{Histogram of predicted classes for 1000 noise images.}
    \label{fig:noise_image_classification}
\end{figure}

\begin{figure}[!ht]
    \centering
    \begin{subfigure}[b]{0.32\textwidth}
        \centering
        \includegraphics[width=1.1\textwidth]{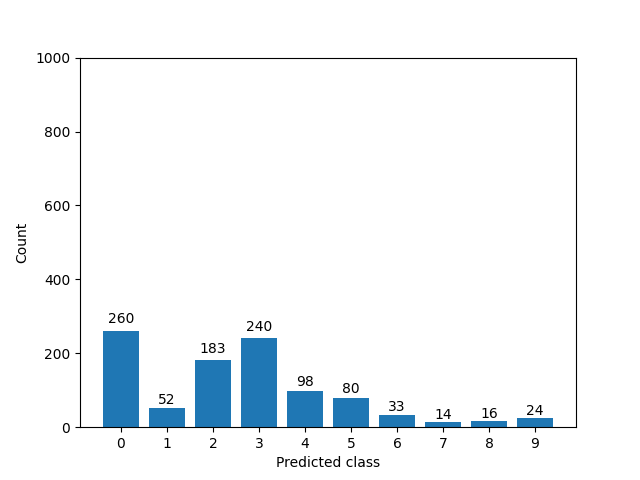}
        \caption{ERM}
        \label{fig:ood_image_classification_erm}
    \end{subfigure}
    \hfill
    \begin{subfigure}[b]{0.32\textwidth}
        \centering
        \includegraphics[width=1.1\textwidth]{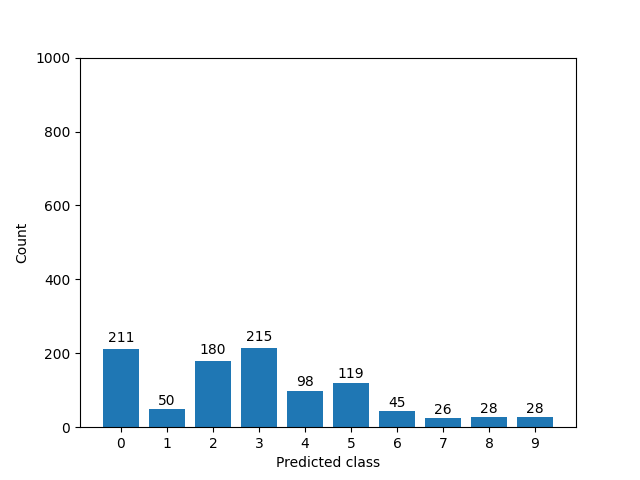}
        \caption{DRS}
        \label{fig:ood_image_classification_drs}
    \end{subfigure}
    \hfill
    \begin{subfigure}[b]{0.32\textwidth}
        \centering
        \includegraphics[width=1.1\textwidth]{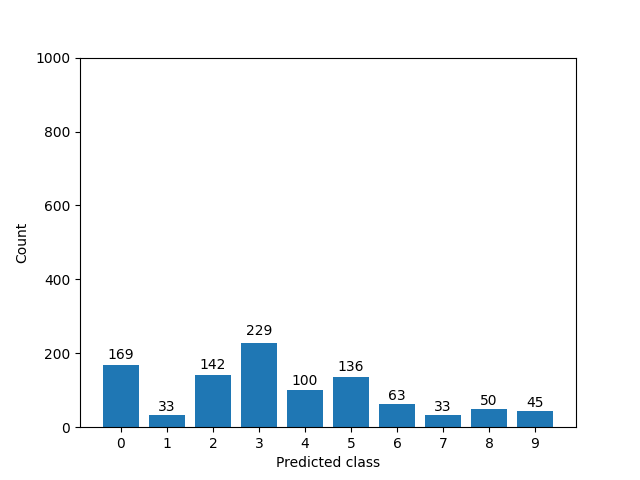}
        \caption{OPeN}
        \label{fig:ood_image_classification_open}
    \end{subfigure}
    \caption{Histogram of predicted classes for 1000 out-of-distribution images.}
    \label{fig:ood_image_classification}
\end{figure}

\end{document}